\documentclass[a4paper,twoside]{article}

\usepackage{bm}
\usepackage{color}
\usepackage{balance}
\usepackage{epsfig}
\usepackage{subcaption}
\usepackage{calc}
\usepackage{amssymb}
\usepackage{amstext}
\usepackage{amsmath}
\usepackage{amsthm}
\usepackage{multicol}
\usepackage{pslatex}
\usepackage{apalike}
\usepackage{wrapfig}
\usepackage{microtype}
\usepackage{textcomp}
\usepackage{multirow}
\usepackage{SCITEPRESS} 

\begin{document}

\title{Identification of Gait Phases with Neural Networks for Smooth Transparent Control of a Lower Limb Exoskeleton }
\author{\authorname{Vittorio Lippi\sup{1}\orcidAuthor{0000-0001-5520-8974}, Cristian Camardella\sup{2}\orcidAuthor{ 0000-0002-3856-5731}, Alessandro Filippeschi\sup{2},\sup{3}\orcidAuthor{0000-0001-6078-6429} and Francesco Porcini\sup{2}\orcidAuthor{0000-0001-9263-9423}}
\affiliation{\sup{1}University Hospital of Freiburg, Neurology, Freiburg, Germany}
\affiliation{\sup{2}Scuola Superiore Sant'Anna, TeCIP Institute, PERCRO Laboratory, Pisa, Italy}
\affiliation{\sup{3}Scuola Superiore Sant'Anna, Department of Excellence in Robotics and AI, Pisa, Italy}
\email{vittorio.lippi@uniklinik-freiburg.de \{cristian.camardella, alessandro.filippeschi, francesco.porcini\}@santannapisa.it }}


\abstract{Lower limbs exoskeletons provide assistance during standing, squatting, and walking. Gait dynamics, in particular, implies a change in the configuration of the device in terms of contact points, actuation, and system dynamics in general. In order to provide a comfortable experience and maximize performance, the exoskeleton should be controlled smoothly and in a transparent way, which means respectively, minimizing the interaction forces with the user and jerky behavior due to transitions between different configurations. A previous study showed that a smooth control of the exoskeleton can be achieved using a gait phase segmentation based on joint kinematics. Such a segmentation system can be implemented as linear regression and should be personalized for the user after a calibration procedure. In this work, a nonlinear segmentation function based on neural networks is implemented and compared with linear regression. An on-line implementation is then proposed and tested with a subject.}

\onecolumn \maketitle \normalsize \setcounter{footnote}{0} \vfill

\section{\uppercase{Introduction}}
\label{sec:introduction}

\begin{figure}[!ht]
\centering
\includegraphics[width=1\columnwidth]{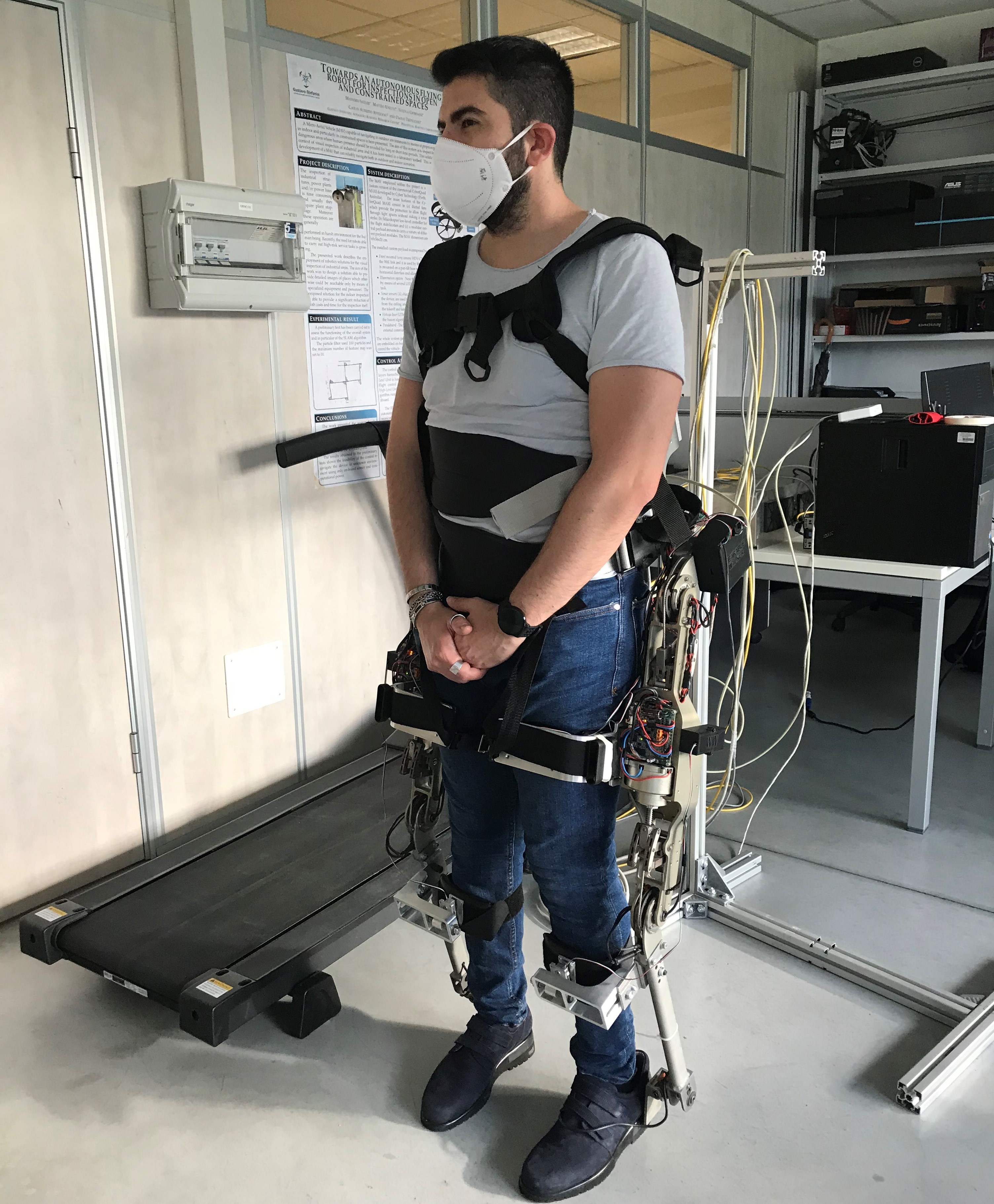}
\caption{The \textit{Wearable Walker} lower limb exoskeleton and the experimental setup used for data collection.}
\label{fig:Fig1Exo} 
\end{figure}

Wearable robots are used to assist the user providing partial compensation for the force required for the performed task. This can find application in a work environment where the user is required to move weight \cite{omoniyi2020farmers,pillai2020evaluation}, in rehabilitation and impairment compensation \cite{afzal2020exoskeleton,swank2020feasibility}. The particular case of medical application requires to address specifically the impairment that affects the patient, but in all these scenarios the robot provides a helping force or torque to the user. The help must be provided coherently with the intention of the user in terms of timing and intensity.
In the specific case of the lower limb exoskeleton considered in this work (Fig. \ref{fig:Fig1Exo}) the helping force consists in a partial compensation of gravity and inertia. As the user walks, the configuration of the body and the robot changes as the supporting leg changes, this requires switching between different controllers. This is achieved by identifying the phase of the gait. Most of the control systems available at the state of the art, e.g \cite{kazerooni2005control,yang2009impedance,yan2015review,tagliamonte2013human,he2007study}, manage the switching between different control schemes, required by different gait phases, using a finite state machine. This produces a discontinuity in assistive torques during phase transitions, such an aspect is not often investigated \cite{yan2015review}. In a previous work, \cite{Camardella2021}, it was shown that the phase can be properly identified with a linear combination of joint angles. Such transformation was obtained through a linear regression performed after a calibration trial (i.e. recording joint angles without controlling the exoskeleton). The aim of the present work consists of the analysis of a phase identification performed using neural networks, using the data from \cite{Camardella2021}. This is motivated by the perspective that machine learning will be an important tool in managing the complexity of human robot interaction (HRI) in assistive devices \cite{Argall2013,broadlearning,aa8311855,aaA.Kurkin.2018,aaNa.2019}. Because the learned estimator of the gait phase is affecting the system itself after the learning through the control system, a slight decrease in performance can be expected. In \cite{lippi2018prediction} experiments with humanoid robots showed that on-line incremental learning is a viable solution for this issue. A test of an online implementation is then tested, with one trial with one subject.

\begin{figure}[t]
\centering
\includegraphics[width=0.7\columnwidth]{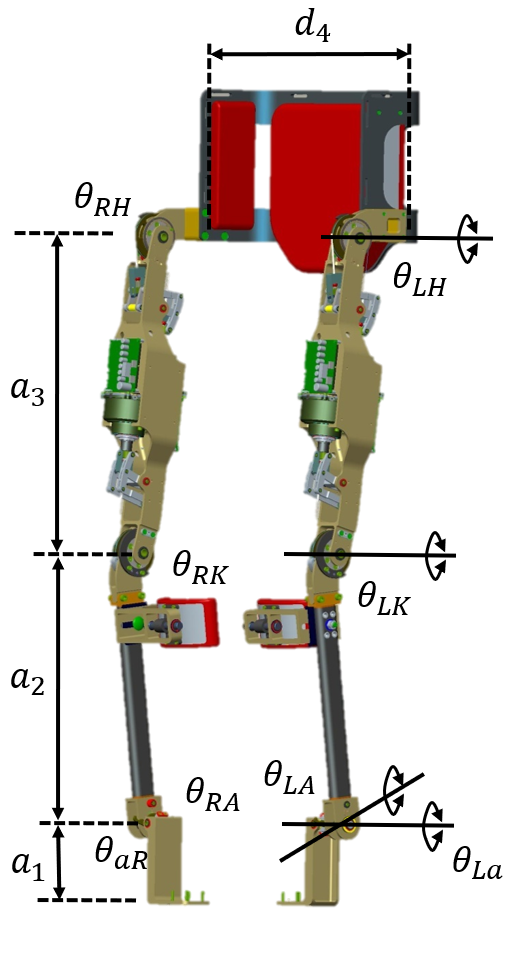}
\caption{The \textit{Wearable Walker} lower limb exoskeleton. The robot has 6 DoFs in the sagittal plane: hips, knees and ankles. Each DoF is equipped with Hall-effect sensors and encoders on motor shafts. There are 2 more passive DoFs to allow ankles inversion-eversion.}
\label{fig:Fig2ExoKin} 
\end{figure}

\section{\uppercase{Materials and methods}}
\begin{figure}[htbp]
\centering
\includegraphics[width=1.00\columnwidth]{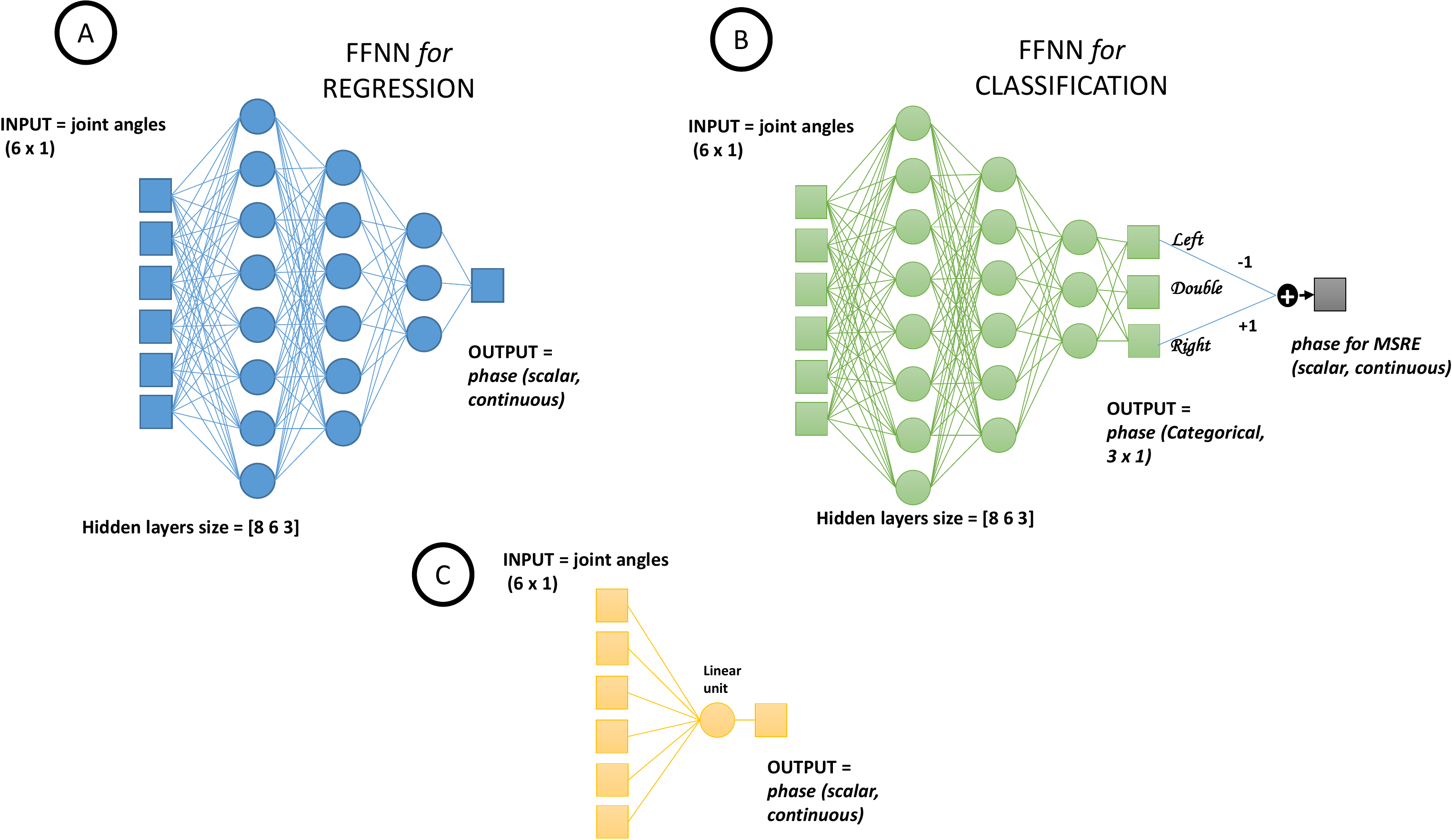}
\caption{Architecture of the neural networks (A,B) and of the linear regression represented as a neural network (C)}
\label{fig:FigureNNarchitecture}
\end{figure}

\subsection{The exoskeleton.} The Wearable-Walker exoskeleton (Fig. 1) is a lower-limb exoskeleton used for assistance in tasks such as load carrying. The exoskeleton is built by a rigid structure of 7 links and 8 revolute joints: hip and knee flexion/extension (active joints) per leg, ankle flexion/extension, and ankle inversion/eversion (passive joints) per leg.
The active joints mount each a brushless torque motor (RoboDrive Servo ILM 70x18) in series to a tendon-driven transmission system composed of a screw and a pulley \cite{bergamasco2011high} guaranteeing a reduction ratio of 1:73.3 and high efficiency (90$\%$). Each joint has been sensorized with Hall-effect sensors, while there are encoders on each motor shaft. Moreover, there are 4 pressure sensors in
each shoe insole and 4 load cells on each human-robot
interface (at the shins and thighs) to monitor human-robot
interactions. The power electronics group consists of a battery of 75 V, 14 Ah, and a voltage conversion board, while the control group consists of a computer running Simulink Real-Time at 5 kHz and a communication module; both groups are housed on the backlink of the exoskeleton. The communication module allows Wi-Fi communication with a host PC (which remotely controls the on-board computer) and the communication via EtherCAT protocol with the low-level control module of each leg. 


\begin{table}[b!]
\begin{tabular}{|c|c|c|c|c|c|}
\hline
\multicolumn{2}{|c|}{Lengths [$mm$]} & \multicolumn{2}{|c|}{Masses [$kg$]} & \multicolumn{2}{|c|}{ROMs [$rad$]} \\
\hline
$a_1$ & 95 & $m_1$ & 0.2 & $\theta_{*A}$ & -70,70 \\
$a_2$ & 402 & $m_2$ & 2.9 & $\theta_{*K}$ & -5,125 \\
$a_3$ & 407 & $m_3$ & 4.1 & $\theta_{*H}$ & -30,125 \\
$a_4$ & 474 & $m_4$ & 1.2 & $\theta_{*a}$ & -50,50 \\
\hline
\end{tabular}
\caption{Lengths [$mm$], masses [$kg$] and joint ranges of motion (ROMs) [$deg$] of the exoskeleton. The ROMs are the same for the two legs.}
\label{tab:masslength}
\end{table}

\subsection{Control strategies.} Identification of gait phases was tested with three different control strategies. Since each control strategy affects the way each subject walks while wearing the exoskeleton, the proposed neural network performance was investigated offline on a previously acquired dataset in which the subjects had active control strategies while walking. The analyzed strategies were FSM (Finite State Machine), sFSM (smoothed FSM), and Blend. All methods apply a feed-forward gravity and inertial compensation, exploiting two dynamic models named LGF (Left Grounded Foot) and RGF (Right Grounded Foot). These models differ in considering a serial chain starting with a base frame either on the left ankle or right ankle. LGF and RGF models are used respectively during a left single stance (swinging right leg) and a right single stance (swinging left leg). The way these models contribute in computing the aforementioned compensation torques characterizes the three analyzed strategies:

\begin{itemize}
\item A. FSM: the control output of this strategy instantaneously switch between LGF and RGF outputs whenever a left or right single stance is detected;
\item B. sFSM: the control output applies a smoothing function with a fixed time constant, when switching from left single stance to right single stance \cite{hyun2017development};
\item C. Blend: the control output depends on two coefficients that, varying continuously with the variations of the joint angle, weighs the influence of LGF and RGF models output. The total output is the sum of the single weighted outputs \cite{Camardella2021}.
\end{itemize}

\subsection{The dataset.} The presented analysis is based on the data from \cite{Camardella2021}. Specifically, the test was performed on 15 subjects, with age $32.45 \pm 5.34$ years and height $1.76\pm 0.04$ (i.e. the 11 subjects who participated in \cite{Camardella2021} and 4 more that were recorded similarly) . All the participants signed a written disclosure and voluntarily joined the experiments. All the experiments have been conducted under the World Medical Association Declaration of Helsinki guidelines. Each subject performed a calibration trial used for the regression wearing the exoskeleton, but with no active control. The calibration trial was performed with a series of different speeds, each one applied on the treadmill for 30 s, specifically 1,2,3,3.5,3,2,1 Km/h. The transition between speeds lasted about 2 seconds. A total of 9 test trials for each subject were performed as follows. Three different control systems were then tested: FSM, SmoothFSM, and Blend. For each control system, three speeds were tested: walking on the treadmill at 1 km/h for 30 s, at 3.5 km/h for 30 s, ground walking for 7 meters with no imposed speed (``free'' in Table \ref{OnlineLin}). 
Since 
The phase is identified using FSR (force sensing resistor) sensors in the soles (see Fig \ref{fig:Fig1Exo}. It should be noticed that while FSR sensors, recognizing the contact with the ground directly, can provide ground truth for the training, in the long run, they require recalibration. This explains why it is needed to base phase recognition on kinematic data (encoders) instead of using the FSR directly. The sampling rate is 100 Hz, the number of samples used for training varied for the different subjects, with a range from 21041 to 11541 with an average of 14125.
\subsection{Regression.} The linear regression was performed between the 6 joint angles and the phase variable, producing a $1 \times 6$ matrix of coefficients. The phase variable is defined as:
\begin{equation}
\varphi = 
\left\{
\begin{aligned}
-1 & &left \\
0 & & double \\
1 & & right \\
\end{aligned} \right.
\label{phaseeq}
\end{equation}
A neural network with the architecture shown in Fig. \ref{fig:FigureNNarchitecture} ``A'' was trained to perform the same regression. A neural network classifier has been trained for comparison (Fig. \ref{fig:FigureNNarchitecture} ``B'' ). The classifier interpreted the 3 possible values of the target phase as a class. Both the neural networks were trained with the Levenberg-Marquardt \cite{hagan1994training}, validation stop was used as regularization principle. The size of the hidden layers was the same for both networks: empirically increasing the size did not produce an observable increase in performance.
\subsection{Neural Network On-line Training.} 
\label{online}
The training of the neural network is performed also with an incremental procedure, working \textit{on-line} in two conditions while the user is walking: with and without the active modality of the robot control. The subject was asked to wear the Wearable Walker exoskeleton and walk on the threadmill at 3.5 km/h for 5 minutes without the assistance and 5 minutes with Blend-control assistance activated. In order to implement such a training in the Matlab environment, an instance of Matlab\textregistered was running in parallel with the compiled control system, acquiring samples via UDP and producing the updated weights for the neural network. The compiled control system contains the neural network that actually produce the estimated gait phase and receives all the weights and biases every time a training cycle is accomplished, after that, the net is updated. The output of the neural network is never used to modify the control assistance logic. Matlab\textregistered's \textit{Statistics and Machine Learning Toolbox\textregistered} does not take in account explicitly the possibility to train a network on-line. In order to implement on-line training manually, the training function \texttt{trains} is used, performing a sequential training on the sample sets. 
The network is set up as: \\
\texttt{ net = feedforwardnet([8 6 3],'trains'); \\
net.trainParam.epochs=1;} \\
The parameter \texttt{trainParam.epochs} set to one means that the batch of samples, received by the second Matlab instance via UDP are used for the training only once. This was a fast implementation, thought to get a result that could be compared with the offline training in the Matlab environment. At the state of the art there are several specific studies on incremental learning, i.e. \cite{icinco19b,icinco19cecca,icnc09,icnc09b}, in future works the presented learning system can be improved. Nevertheless, at the time of writing this paper, the issue of incremental training implemented in {Matlab\textregistered}  appeared to be of interest for the users' community (i.e. a topic of several forums), we therefore assume that the implementation presented herein may be useful as an example for the readers. The neural network was initialized with small random weights when both models (control and neural network training schemes) started, and was scheduled to be trained every 20 seconds. The training function was called through a Simulink block 'PostOutput' listener that was also delegated to fill a circular buffer with input and output data. When the scheduled training time was reached, the collected input and output data fed the \texttt{trains} training function. The dimension of the circular buffer was set in a way that it was already filled in the instant of the first training cycle. After each training, weights and biases of the neural network were retrieved and sent to the compiled control scheme. If the neural network was busy in the scheduled training step, that cycle was postponed 20 seconds after.

\section{\uppercase{Results}}
\subsection{Regression and classification}
\begin{figure*}[htb]
\centering
\includegraphics[width=1.00\textwidth]{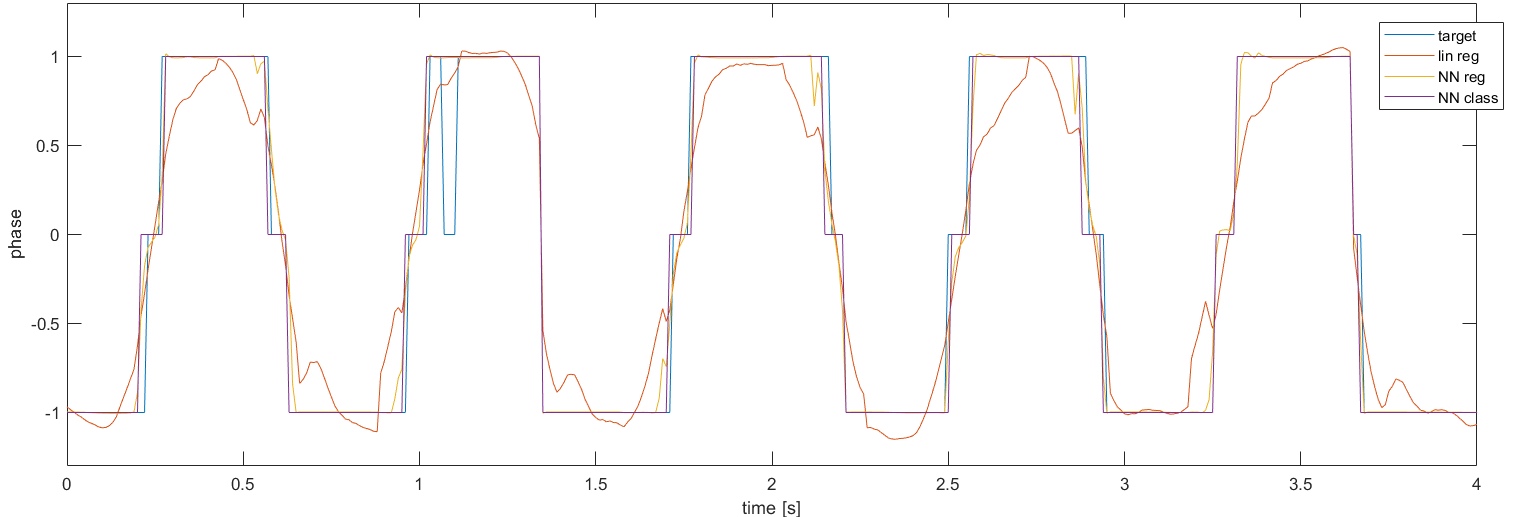}
\caption{Example of output for the linear regression \textit{lin reg}, the neural network regression \textit{NN reg} and the neural network classifier \textit{NN class}. The output is plotted together with the phase used for training \textit{target}. Notice the irregularity in the original data around 1 s, such an event contribute substantially to the RMSE, especially for the neural networks, the output of which is closer to 1.}
\label{timeFit}
\end{figure*}

In order to evaluate the performance of the different regression methods, 5-fold cross-validation has been performed. Specifically the dataset was split into 5 parts of equal duration in time. The regression was performed using a calibration trial for each subject. The performance is computed for each subject separately and the final result is computed as the average of all the results for the subjects. The results obtained with FFNN and linear regression are compared in Table \ref{tab:my-table}. The nonlinear model provided by the neural network produced a better fit (smaller RMSE) than the linear regression, associated with an increase of performance in terms of recognizing the phase. The classification was implemented considering a threshold applied to the output of the regression and of the neural network. 
\begin{equation}
\varphi(y_{reg}) = 
\left\{
\begin{aligned}
-1 & & y_{reg} < -\theta \\
0 & & \theta > y_{reg} > -\theta \\
1 & & y_{reg} > \theta \\
\end{aligned} \right.
\end{equation}

where $y_{reg}$ is the output of the regression (or FFNN) the value $\theta =0.1$ was decided empirically in \cite{Camardella2021}. For comparison, a neural network with similar architecture to the one used for the fit, but designed and trained for classification is also analyzed (see Fig. \ref{fig:FigureNNarchitecture} ``B''). Such a network outputs the class. In order to compute the accuracy, the categorical output was translated into a numerical value $\varphi$ following the definition of eq. \ref{phaseeq}. The classifier network produced a better accuracy but a worse RMSE. This is not surprising considering that with the output defined on the domain $\{ -1,0,1 \}$, each classification error produced a quadratic deviation (i.e. 1 or 4) that was larger on average in comparison to the errors obtained with the continuous variables from both the linear and FFNN-based regressions; an example is shown in Fig. \ref{timeFit}.
\begin{table}[ht]
\center
\begin{tabular}{|l|l|l|}
\hline
& RMSE & Accuracy\\ \hline \hline
Linear & 0.51 & 82.56 \% \\ \hline
FFNN fit & 0.37 & 84.57 \% \\ \hline
FFNN class & 0.75 & 87.04 \% \\ \hline
\end{tabular}
\caption{Average prediction error and classification accuracy obtained with linear regression and neural networks. The calibration data are used as training and test set with a 5-fold cross validation. The result across the 15 subjects is then averaged.}
\label{tab:my-table}
\end{table}

\begin{figure}[htbp]
\centering
\includegraphics[width=1.00\columnwidth]{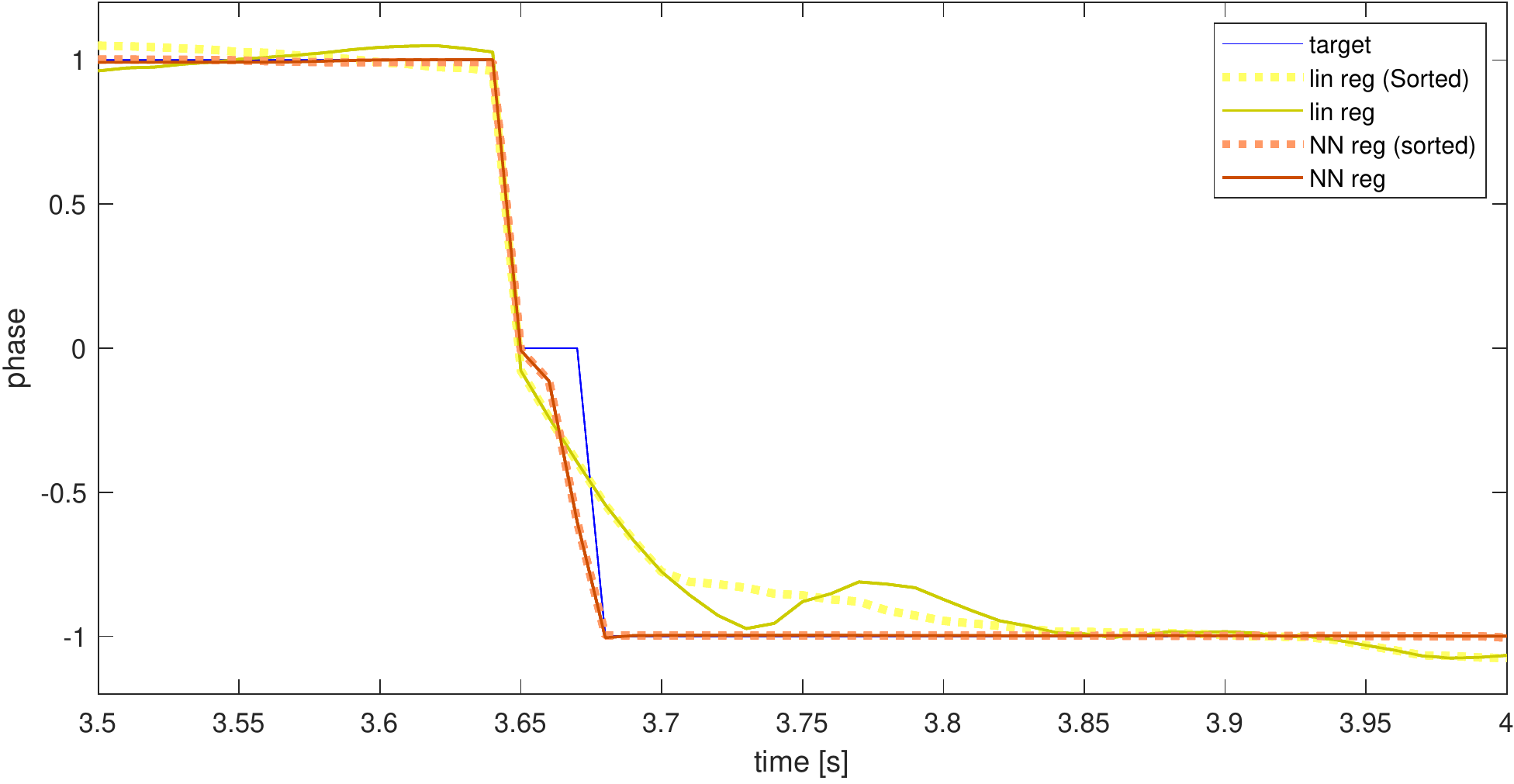}
\caption{The output of the NN is \textit{more monotonic} in time compared with the one of the linear regression. This is quantified comparing the samples in chronological order with the samples sorted by decreasing magnitude over approximately a quarter of cycle. the RMSE between sorted and unsorted is $0.0491$ for the linear fit and $0.0042$ for the NN.}
\label{monotone}
\end{figure}

A characteristic  response of the linear regressor is the presence of spikes in the signal, this is visible in Fig. \ref{timeFit}. The phenomenon is happening sometimes also for the NN, e.g. Fig. \ref{timeFit} around $3 s$ but with a smaller amplitude and not for each cycle. In order to quantify it Fig. \ref{monotone} shows the signals of Fig. \ref{timeFit} between the $2.75 s$ and $3.2 s$. The monotonicity of the function is evaluated by computing the RMSE between the measured signal and a version where the samples are sorted, it accounts on average to $0.0491$ for the linear regressor and $0.0042$ for the neural network respective outputs (similar values are measured for the rest of the dataset). This provides a smoother transition.

\begin{table}[ht]
\center
\begin{tabular}{|l|c|c|c|}
\hline
& Speed & RMSE & Accuracy\\ 
& [Km/h]& & \\ \hline \hline
Training & $0\leftrightarrow 3.5 $& 0.44* & 83.51* \% \\ \hline
FSM & 1 & 0.43 & 70.23 \% \\ \hline
FSM & 3.5 & 0.43 & 79.41 \% \\ \hline
FSM & free& 0.53 & 73.49 \% \\ \hline
SFSM & 1 & 0.43 & 76.46 \% \\ \hline
SFSM & 3.5 & 0.43 & 78.65 \% \\ \hline
SFSM & free& 0.56 & 69.70 \% \\ \hline
Blend & 1 & 0.42 & 74.28 \% \\ \hline
Blend & 3.5 & 0.42 & 77.89 \% \\ \hline
Blend & free& 0.53 & 67.06 \% \\ \hline
\end{tabular}
\caption{Performance obtained by the linear regression once the phase recognition is used in ``closed loop'' when controlling the exoskeleton with three different control systems. The results is averaged over the 15 subjects. The training result is different from the one reported in Table \ref{tab:my-table} because in this case it was computed on the whole dataset without cross validation.}
\label{OnlineLin}
\end{table}
\subsection{Neural Network On-line training}
\begin{figure}
	\centering
		\includegraphics[width=1.00\columnwidth]{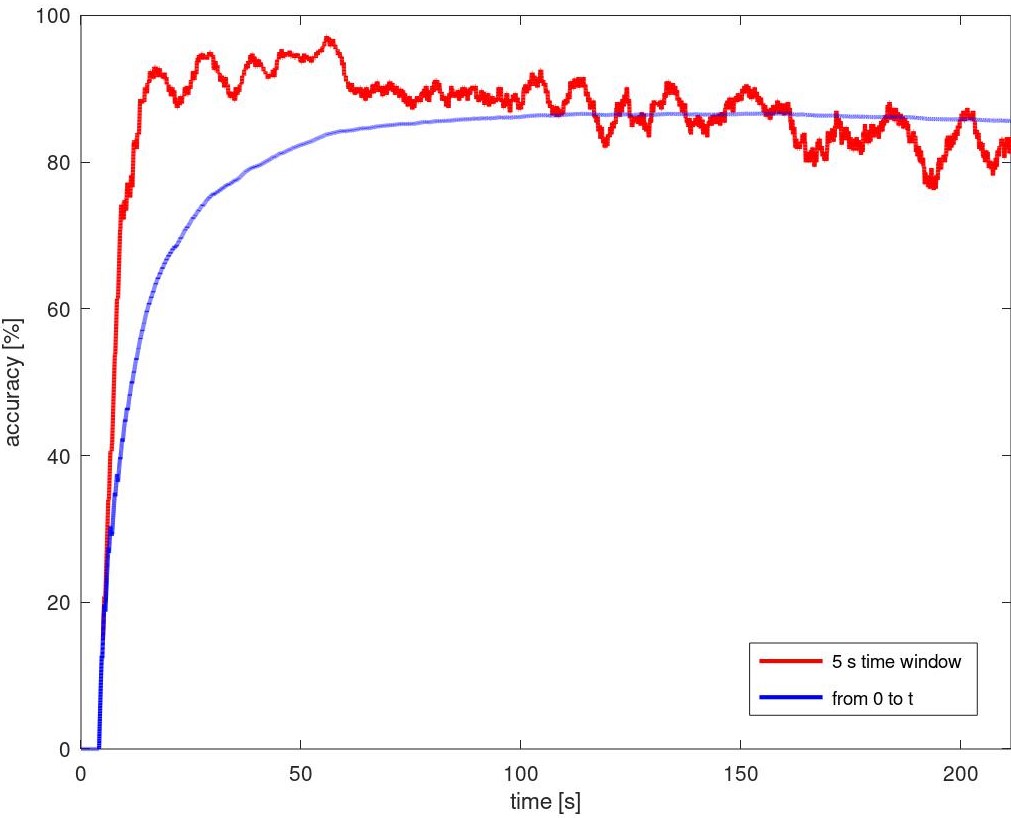}
	\caption{Accuracy of the real-time neural network. The red line represents the accuracy computed on a time window spanning the last 5 seconds, the blue line represents the accuracy over the whole dataset up to the current time.}
	
		\label{fig:PerformanceTEMP}
\end{figure}

\begin{figure}
	\centering
		\includegraphics[width=1.00\columnwidth]{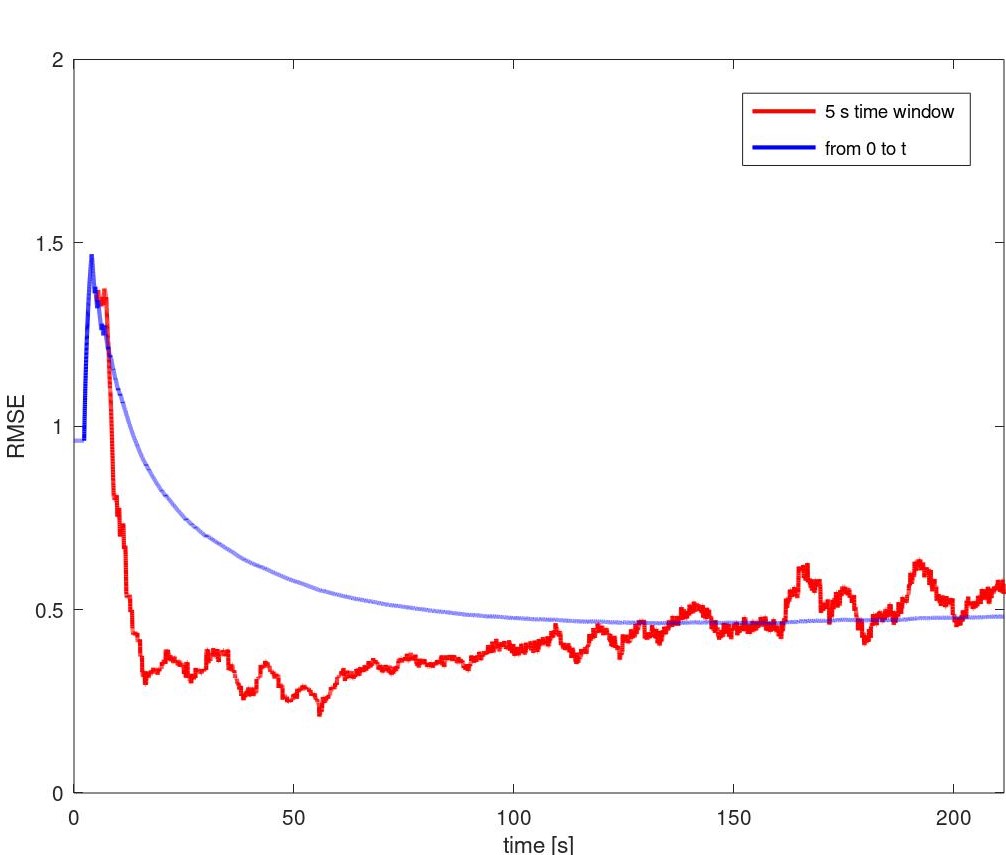}
	\caption{RMSE of the real time neural network. The red line represents the RMSE computed on a time window spanning the last 5 seconds, the blue line represents the RMSE over the whole dataset up to the current time.}
	
		\label{fig:RMSETEMP}
\end{figure}
The results of the on-line training test with the control active, described in \S \ref{online} are shown in Fig.\ref{fig:PerformanceTEMP} and Fig. \ref{fig:RMSETEMP}, displaying accuracy and RMSE respectively. The accuracy was overall better than the one observed in the off-line cases ($\leq$ 80 \% ). The use of a 5-seconds time window showed peaks of performance locally that are not affected by the initial bad performance. At the same time a slight decrease of the performance at the end of the trial, on average the value on the time window after the first 10 seconds was 87 \%. Similar considerations can be done for the evolution of \ref{fig:RMSETEMP} in time. It should be noticed that, notwithstanding the improvement in the accuracy, the RMSE is slightly larger than the one obtained off-line.    

\balance
\section{\uppercase{Discussion, Conclusions and future work}}
In this work we analyzed the exoskeleton control presented in \cite{Camardella2021} from the point of view of the recognition of gait phase using joint angles, proposing and testing an approach based on neural networks. Neural networks perform better than linear regression, yet the linear regression is not significantly worse. Both solutions seem useful in this context. A further improvement in classification accuracy can be obtained by training a classifier explicitly instead of a regression associated with a threshold (similarly to the classifier proposed in the offline case). Nevertheless, it should be noticed that in this context the estimated $\varphi$ is used to control smooth transitions between control systems. This means that a function that varies slowly between the three target values $-1$, $0$ and $1$ can be preferable to a function that changes more abruptly pursuing smaller RMSE respect to the reference, considering that such reference moves in steps as shown in Fig. \ref{timeFit}. Future work will exploit the versatility of the neural network to optimize the transition dynamics explicitly. In this sense, online training can be used to reduce the interaction forces between the user and the robot during the transitions, an objective function that is available only when the control is active. It should be noticed that the transitions are not the only issue that can increase interaction forces: for example, the double stance phase represents a hyper-static configuration where the relative contribution of the two legs is arbitrary \cite{goodworth2012sensorimotor}. In humanoid control, additional constraints may be added in the form of synergies\cite{hauser2007biologically,hauser2011biologically,alexandrov2017human,lippi2016human}, but this may not reflect the natural human behavior: it is known that the weight-bearing is asymmetric both in healthy subjects and in patients. Such issue should be addressed in the design of the double stance controller, even taking into consideration that the two legs provide independent sensory feedback, that could be an impact on how the system is perceived. This specific study focused on the identification of the gait phase, the long term project aims to a more general evaluation of performance in gait and balance: the formal definition of benchmarking performance indicators is currently under research for gait \cite{torricelli2020benchmarking} and balance \cite{Lippi2019,lippi2020performance}.
The test of the on-line learning showed a stable convergence to a trained network that had a performance that was better than the one of the network trained off-line in terms of accuracy. This suggests a possible advantage in using on-line learning for exoskeletons when the neural network is used in the control loop, similarly to what is observed with humanoid robots \cite{lippi2018prediction}. The convergence speed of the on-line training is limited by the design choice of using each new batch of training data only of 1 epoch. In future work more ''aggressive´´ training systems will be tested (e.g. using batch of old and new data together and iterate for more epochs). The test proposed in this work was focused on the global result, future studies will also take in account how human sensor fusion interacts with the proposed control\cite{Hettich2013,hettich2015human} also considering the subjective report of the subjects \cite{luger2019subjective,stampacchia2016walking}.
\newpage
\section*{\uppercase{Acknowledgements}}
This work is supported by the project EXOSMOOTH, a sub-project of EUROBENCH (European Robotic Framework for Bipedal Locomotion Benchmarking,
\begin{wrapfigure}{l}{0.22\columnwidth}
\vspace{-12pt}
{\includegraphics[width=0.3\columnwidth]{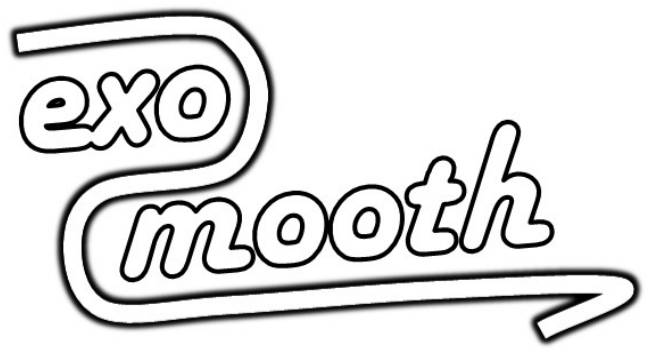}}
\vspace{-20pt}
\label{LOGO}
\end{wrapfigure}
\noindent www.eurobench2020.eu) funded by H2020 Topic ICT 27-2017 under grant agreement number 779963.

\bibliographystyle{apalike}
{\small
\bibliography{example}}

\end{document}